\documentclass[letter,10pt,conference]{resources/ieeeconf}
\IEEEoverridecommandlockouts
\usepackage{amsmath,amssymb,theorem}
\usepackage{graphicx}
\usepackage{hyperref}
\usepackage{algorithm,algpseudocodex}
\usepackage{psfrag}
\usepackage{mathtools}
\usepackage{siunitx}

\usepackage{color,xspace}
\usepackage{subfigure}

\usepackage{etoolbox,environ}
\newtoggle{archive}
\toggletrue{archive}
\NewEnviron{archiveText}
  {\iftoggle{archive}{\BODY}{}}

\newcommand{\real}{{\mathbb{R}}}

\renewcommand{\epsilon}{\varepsilon}

\renewcommand{\hat}{\widehat}

\newcommand{\oprocendsymbol}{\hbox{$\bullet$}}
\newcommand{\oprocend}{\relax\ifmmode\else\unskip\hfill\fi\oprocendsymbol}

\renewcommand{\theenumi}{(\roman{enumi})}

\newcommand{\erroryaw}{e_{\text{yaw}}}
\newcommand{\errorvert}{e_{\text{vert}}}
\newcommand{\ballincage}{\gamma}
\newcommand{\ballinsight}{\sigma}
\newcommand{\goalinsight}{\chi}
\newcommand{\famode}{\zeta}
\newcommand{\fwdvelsel}{\bar{u}_1}
\newcommand{\turnvelsel}{\bar{u}_2}
\newcommand{\updwnvelsel}{\bar{u}_3}

\begin{document}

\title{Lighter-Than-Air Autonomous Ball Capture and Scoring Robot \\ Design, Development, and Deployment}

\author{Joseph Prince Mathew \quad Dinesh Karri \quad James Yang \quad Kevin Zhu \quad Yojan Gautam   \\  Kentaro Nojima-Schmunk \quad Daigo Shishika \quad Ningshi Yao \quad Cameron Nowzari \thanks{All authors are with the Electrical and Computer Engineering Department, George Mason University, Fairfax, VA 22030, \{jprincem, cnowzari\}@gmu.edu. 
    }}
  
\maketitle

\begin{abstract}
This paper describes the full end-to-end design of our primary scoring agent in an aerial autonomous robotics competition from April 2023. As open-ended robotics competitions become more popular, we wish to begin documenting successful team designs and approaches. The intended audience of this paper is not only any future or potential participant in this particular national Defend The Republic (DTR) competition, but rather anyone thinking about designing their first robot or system to be entered in a competition with clear goals. Future DTR participants can and should either build on the ideas here, or find new alternate strategies that can defeat the most successful design last time. For non-DTR participants but students interested in robotics competitions, identifying the minimum viable system needed to be competitive is still important in helping manage time and prioritizing tasks that are crucial to competition success first. 
\end{abstract}

\section{Introduction: Defend The Republic}\label{se:dtr}
% Robotics competitions \cite{robocupRoboCupFederation, vexroboticsCompetitionRobotics, robooneBipedRobot} are seeing increasing uses for promote and advance both education and research in STEM. 

\begin{figure}[t]
    \centering
    \includegraphics[width=\linewidth]{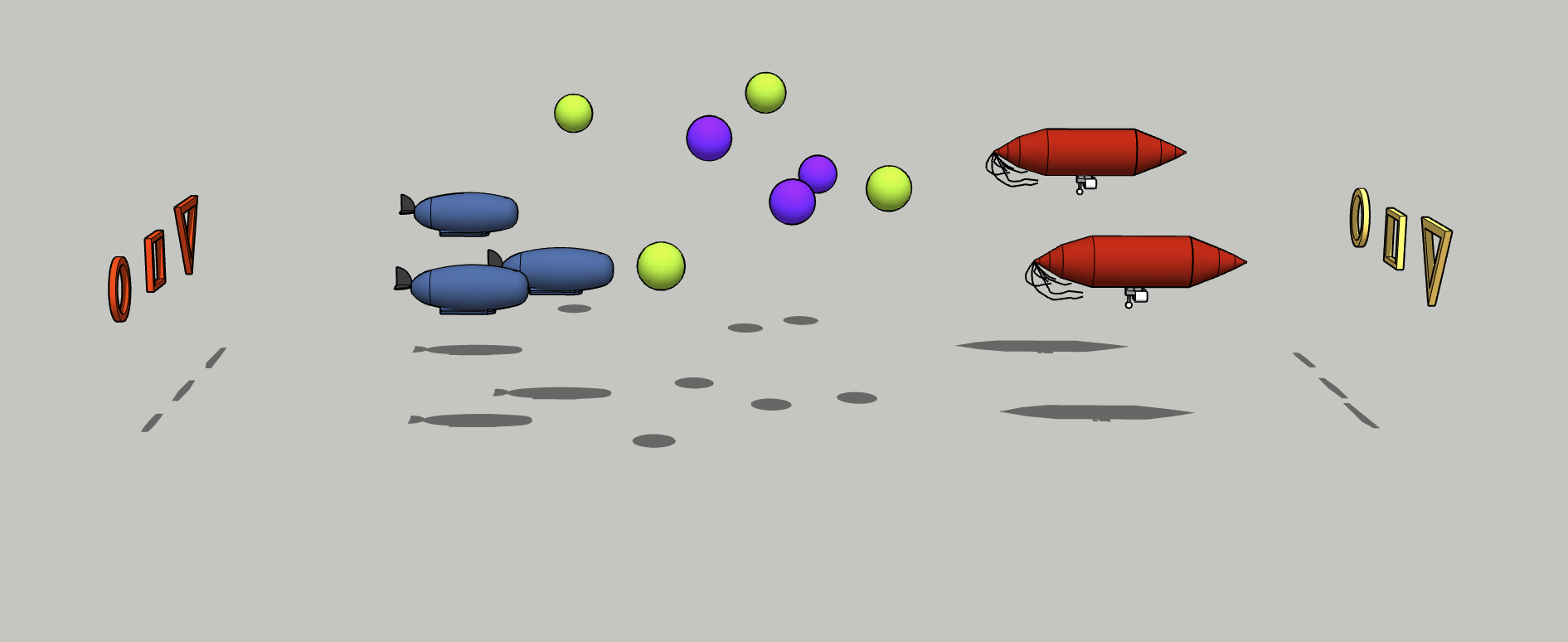}
    \caption{Defend The Republic (DTR) game scenario showing two teams (Red versus Blue) with green and purple balls scattered in the environment and goals (one triangular, circular and square per side) in fixed locations. The agents must capture these green balls and score by completely passing the ball through the suspended goals.}
    \label{fig:blimps_scenario}
\end{figure}

Defend The Republic (DTR) is a national Lighter-Than-Air (LTA) robotics competition that pits two teams (Red versus Blue) against each other in a 60-minute head-to-head match in which fleets of autonomous robots must capture Green and Purple neutrally buoyant balls and move them through Circle, Square, and Triangle goals suspended from the ceiling. An overview of the game with all the elements are shown in Fig.~\ref{fig:blimps_scenario}. 
The environment being shape- and color-coded allows easier perception so the teams can focus on advancing multi-agent control and interaction problems. 

%
%n order to focus on the multi-agent control and interactions problem, the game is color-coded to ease the level of required perception. Teams are required to have red and blue fleets, game balls are either green or purple, and the goals are yellow and orange as shown in Figure~\ref{fig:blimps_scenario}. In this paper we discuss the design and development of the winning agent that was deployed in the latest game of DTR working within the rules mentioned above.
%Before participating in the head-to-head matches, teams must first qualify by scoring a goal within 30 minutes of unopposed gameplay. Once qualified, teams engage in the main game, which involves the use of multiple game balls. These game balls are designed to be neutrally buoyant and float around the arena randomly. The teams defend three hoops mounted from the ceiling in their respective end zones, each hoop featuring different shapes and coated with retro-reflective tape. Goals are scored by moving game balls through the opponent's goals. Each full cycle autonomous goal secures the team a total of 10 points for each goal scored and 1 point for green ball captured.

%, with specific point values assigned to different scoring methods.

The complete rules of the game allow flexibility for a myriad of different control and game strategies including heterogeneous teams (where completely  different agents or robots are used cooperatively like a defending robot passing a ball to an attacking robot that is perhaps faster and more suitable for finding and scoring goals). However, this paper focuses on a single minimally viable agent design and deploying a homogeneous team of them to autonomously play DTR. It is expected that future competitions will see specialized agents that can perform certain tasks better than others as the competition evolves.

%\subsection*{Minimum Viable System}
In considering the design and deployment of a new robot to enter a competition in less than one year, it is critical to identify the most critical aspects of the design and build a full minimum viable system as soon as possible. This allows the team to become competitive as quickly as possible and once a base functional design is found, future iterations need only to start building on previously tried and tested methods. 

The minimum capabilities required for a single LTA agent to successfully capture a ball and move it through a goal are a basic level of aerial locomotion, perception, and interaction with the environment in the form of a method to move around neutrally buoyant balls.
 %For simplicity, we focus here on the design of an agent using only a single camera for visual data but discuss ways the system could be augmented.

It should be emphasized that the evolving design of this agent has been a very iterative process with lots of trial and error as with any robotics competitions.
More specifically, while we identify the critical areas and problems that need solutions as fast as possible, we only present our solution at the time of the April 2023 competition and do not discuss earlier implementation ideas in detail. We will comment on a few earlier design choices to contrast certain earlier choices, we do not discuss any design decisions and instead focus on George Mason University's combination of technologies that happened to be successful in April 2023. 

%The hope is that this paper will immediately highlight and motivate the need for various real-world implementations of state-of-the-art research and other untested technologies demonstrated only in simulations, to truly advance the state of the art. 

In Section~\ref{se:hardware} we identify the minimum viable components necessary to perform all the tasks needed to capture and score a ball in a game of DTR. 

\begin{enumerate}
\item Move around in 3D space for 30 minutes (one half);
\item Find and capture green/purple circular objects;
\item Find orange/yellow, square/circle/triangle-shaped goals and release the ball through it.
\end{enumerate}

In Section~\ref{se:model} we formalize the sequence of tasks our robots need to complete in a conceptual model and present our software that ties all the hardware together.

\section{Minimum Viable System and Hardware}\label{se:hardware}
We partition the required hardware into four components: 

The \textbf{sensors} provide all the raw information available for perception and control.		

The \textbf{actuators} provide all the mechanisms for the agent to move and interact with the world. 

The \textbf{envelope} is the main helium-filled containment or balloon providing nearly all the lift to the entire agent. 

The \textbf{gondola} is the main structure that maintains the integrity of the entire agent and potentially houses any required electronics.

A major challenge of most robotic competitions is the seamless integration of all subsystems actually working together, at the same time. The most sophisticated capturing and scoring robot cannot score a single goal if its envelope isn't large enough to support the helium needed to lift the robot into the air. Being an LTA robot competition, a significant challenge of our design problem is weight. The $200 \unit{cu.ft}$ of helium limit per team means the heavier a single robot is, the fewer robots we can have playing on our team. The total weight of components used for the gondola, sensors, and actuators must all be supported by an envelope large enough to provide the required lift. %In the spirit of supporting our research group's swarming research projects, our design goal was to create the largest number of minimally viable (and thus lightest) agents capable of just barely playing the game. 
%We now present our final design that was used in April 2023. 

%\begin{figure}
%    \centering
%    \includegraphics[width=\linewidth]{images/Weight2nagents.png}
%    \caption{Max Number of agents allowed against the weight of a single agent. With our agents coming to about $621\unit{g}$, we can have a maximum of 4 agents}
%    \label{fig:weight2nagents}
%\end{figure}

\subsection{Sensing}
The color-coded game encourages visual data as a primary way of making sense of the environment. In order to be able to identify different shapes and colors in a large environment, an RGB camera is the natural first choice of sensor. 
In our case we use a monocular USB camera (OV5640) mounted directly at the front of our robot. In terms of minimum viability, no other perception is needed (or used) in our design. Again it should be noted that it is expected future designs can integrate other sensing mechanisms to further improve individual agent performance and ultimately overall team performance.

For instance, a past design included a separate single point LIDAR sensor used solely for detecting whether the robot was currently `holding' a game ball or not. Although this was useful and successful, the added weight was not worth the minor improvements in performance as the single camera is still able to help determine whether a ball is currently being held or not, albeit not as reliably. 

%To detect a ball capture, we use a single Point LIDAR sensor that is mounted in the front of the blimp. When a ball enters the capture cage, the change in the distance reading signals the capture event. 

\subsection{Actuation}
\begin{figure}[!ht]
\centerline{\includegraphics[width=0.9\linewidth]{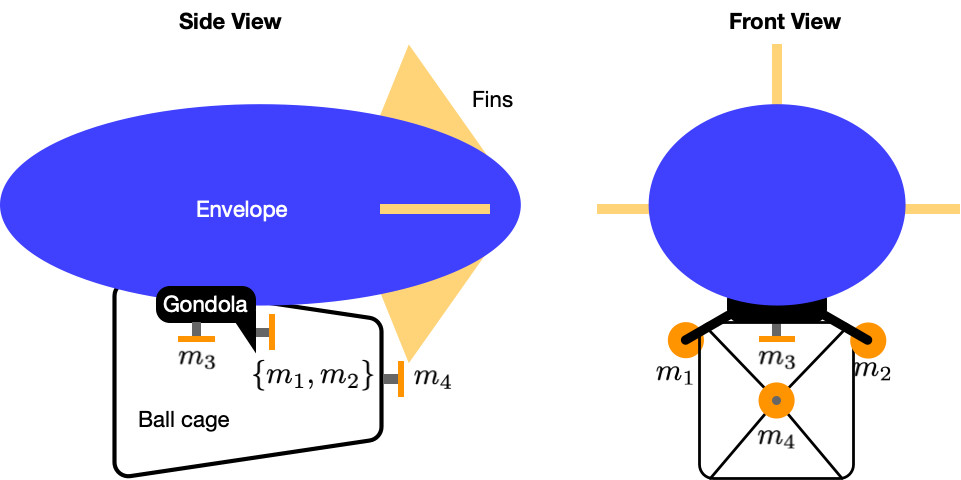}}
\caption{Simplified blimp diagram with motors and cage. }
\label{fig:simpleBlimp}
\end{figure}

To enable our robot to reach any arbitrary point in a 3D environment, we use a 4-motor/propeller combination to allow unicycle-like control (2D position and orientation) coupled with a propeller to allow vertical motion similar to~ \cite{Ferdous2019, Burri2013}. Fig.~\ref{fig:simpleBlimp} shows the motor configuration. 

To control the yaw of the robot, a differential drive mechanism was created using a pair of motors and propellers~$m_1,m_2$ placed 1000mm apart.  Altitude/height control is performed using a third motor/propeller combo~$m_3$ mounted at the bottom of the robot that can produce vertical thrust. Although the above three motors are sufficient for minimally endowing the agent with its required capabilities, we are able to simplify our software problem formalized in Section~\ref{se:problem} by adding a fourth motor~$m_4$ to the back of the robot as the primary forward-thrust mechanism. This allows us to decouple the yaw control and thrust control by delegating them to different motors. 
%A simple schematic shows the propeller configuration in Figure~\ref{fig:simpleBlimp}.

In order to effectively capture neutrally buoyant balls, we have one additional brushless DC servo motor that actuates the gate or door of the cage shown hanging under the envelope in Fig.~\ref{fig:simpleBlimp}. The cage is used to capture, hold, and maneuver balls around the environment. Details about the cage design are in~Section~\ref{se:gondola}. The location of~$m_4$ in Fig.~\ref{fig:simpleBlimp} now enables a dual use not only by providing the primary thrust to the agent, but also serving to blow balls out of the cage to score a goal when operated in reverse.

\subsection{Gondola Structure}\label{se:gondola}
The gondola is a mechanical structure on the blimp which holds all the electronics and actuators on the blimp. This is important in maintaining the structural integrity of the robots and ensuring repeatability and uniformity (reducing idiosyncrasy) across our fleet of agents. 
%We previously tried mounting all the components directly on the envelope of the agent. It was found that as the envelope loses helium pressure inside, a floppiness is introduced in mounts which will affect the dynamics of the robot. Hence, a traditional gondola is created. 
We use a $4 \times 4 \times 1000$mm carbon fiber rod and 3D printed components to mount the 4 motors and the processing unit.
Additionally, we have designed a cage that hangs underneath the envelope as shown in Fig.~\ref{fig:simpleBlimp}. The cage is used by the robot to capture balls and maneuver them around. \begin{archiveText} Further details on the cage design are in Appendix~\ref{se:cage}. \end{archiveText}

\subsection{Envelope Design}
The envelope to hold helium that provides the primary lift for the agent as dictated by the competition is made out of metalized film \cite{Lopez2009, Gorjup2020}. In order to have enough lift to support the payload on the agent, a custom envelope used so we can choose its volume and shape. One side of the film is shiny, silver, and uncolored made with Linear Low Density PolyEster (LLDPE), while the other side has the color of our agent (either light blue or red) made with metalized nylon. When the silver sides of the film are in contact and heated, an airtight seal is created. 

\begin{figure}
    \centering
    \includegraphics[width=\linewidth]{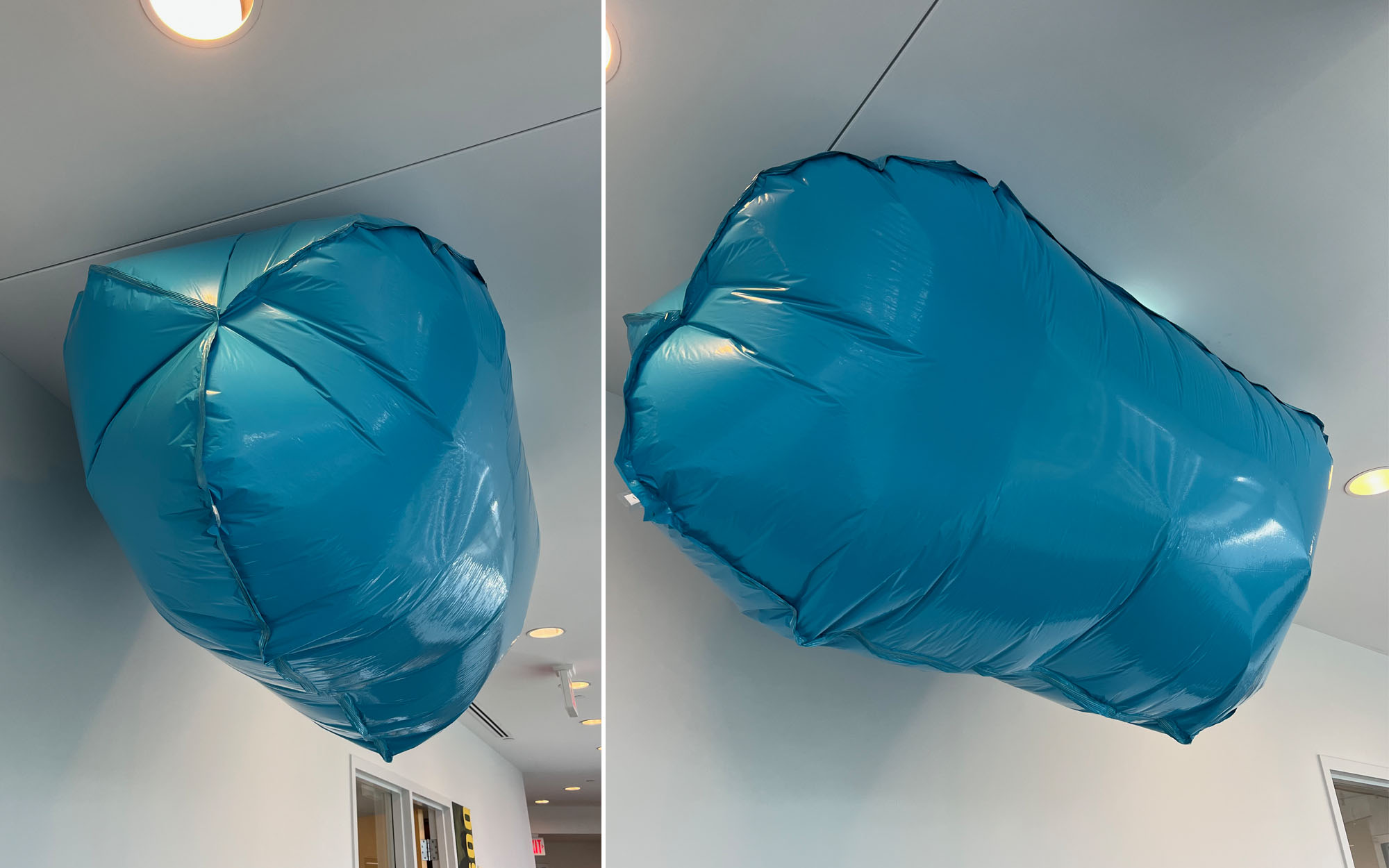}
    \caption{Completed and inflated 3-fold envelope design.}
    \label{fig:Envelope_Completed}
\end{figure}

The total weight of our agent and its respective components used in this agent are shown in Table~\ref{tbl:weights} and came out to $645\unit{g}$. Since helium can in general lift about 1 gram per liter (under `normal' conditions of 25 degrees C and 1 atm pressure) and the game must be played in sometimes varying conditions, we add a 20\% safety factor and created our envelope to hold about~$770$ liters. This ensures we have sufficient lift and can reach any point in 3D space, even in moderately varying environmental conditions. To deal with varying conditions, we use pliable playdough to get our agent as close as possible to neutrally buoyant, while also being able to very easily control the center of gravity of the agent. Fig.~\ref{fig:Envelope_Completed} shows the final envelope used. \begin{archiveText} Details on how these envelopes are designed and produced are in Appendix~\ref{se:envelope}. \end{archiveText}

\begin{table}
    \label{tbl:weights}
    \centering
    \begin{tabular}{|l|l|l|l|}
        \hline
        \textbf{Subsystem} & \textbf{Component} & \textbf{Weight ($\unit{g}$)} & \textbf{Cost (\$)}\\
        \hline \hline
        Sensing     & Camera                    & 15.3      & 35 \\
        \hline
        Actuation   & Motor and Prop.           & 73.3      & 95  \\
                    & Connectors                & 7.2       & 10 \\
                    & ESC                       & 26.6      & 30 \\
                    & Gate servo                & 4.3       & 2 \\
        \hline
        Envelope    & Mylar outer layer         & 154       & 36 \\
        \hline
        Gondola     & Board                     & 71.4      & 380 \\
                    & Board Holder              & 14.8      & - \\
                    & Battery                   & 192.4     & 26 \\
                    & Carbon Fiber Structure    & 82.3      & 30 \\
                    & Adhesive Tape             & 3         & 2 \\
        \hline
                    & \textbf{Total}            & 645       & 646 \\
        \hline
    \end{tabular}
    \caption{Breakdown of individual component weights and costs}
\end{table}

\section{Minimum Viable System Conceptual Model}\label{se:model}

With our basic hardware selected to enable the minimum capabilities required to be viable, we now formalize our problem and discuss our simple 4-mode controller to autonomously play DTR depending on the evolving situation. The intuitive idea is to sequentially move through the following 4 modes of operation:

\begin{enumerate}
\item \textbf{Ball Search. ($\famode=1$)} Simply spin in circles until a green or purple object is found.
\item \textbf{Ball Capture. ($\famode=2$)} Use a tuned PD controller to drive the robot to the found object.
\item \textbf{Goal Search. ($\famode=3$)} Simply spin in circles until the desired orange or yellow object is found. 
\item \textbf{Goal Score. ($\famode=4$)} Use a tuned PD controller to drive the robot to the goal and release the ball.
\end{enumerate}

Clearly, these operations should not always happen in a seamless sequence. For instance if capturing a found ball fails in Step 2 and loses sight of the ball, the agent should go back to Step 1. To simplify perception as much as possible, we rely on a subsumption architecture-like control architecture to create a 4 state finite automata, with higher levels of operation subsuming the lower levels of operation~\cite{Brooks1986, Arkin1998}.

\begin{algorithm}[t]
    \caption{Mode Perception}
    \label{alg:perception}
    \hspace*{\algorithmicindent}  \textbf{Output} current mode~$\famode \in \{1,2,3,4 \}$, target error from center of frame $\erroryaw, \errorvert$ for modes $\famode \in \{2, 4\}$
    \begin{algorithmic}[1]
        \If {$\ballincage = 0$~(not holding ball)}
            \If {$\ballinsight = 0$~(not seeing ball)}
                    \State ~mode $\famode = 1$ (Ball Search)
               \Else 
                    \State ~mode $\famode = 2$ (Ball Capture)
                    \State $\erroryaw =$~horizontal error for ball
                    \State $\errorvert =$~vertical error for ball
               \EndIf
           \Else
               \If {$\goalinsight = 0$~(not seeing goal)}           
                   \State ~mode $\famode = 3$ (Goal Search)
               \Else
                   \State ~mode $\famode = 4$ (Goal Score)
                    \State $\erroryaw =$~horizontal error for goal
                    \State $\errorvert =$~vertical error for goal
           \EndIf
           \EndIf
    \end{algorithmic}
\end{algorithm}

Thus rather than thinking about the tasks as an always sequential operation, we instead enable simple perception to determine which behavior should be driving the robot at any given time. We describe this architecture in levels where the highest level controller should always take over the lower level ones when they are able to. 

\textbf{Level 1: Search.} When no target is available to the robot, it should wander around until it finds something of interest. 

\textbf{Level 2: Go To.} When a target of interest is found, it should go towards it.

\textbf{Level 3: Capture.} If the target is a ball, the robot should capture it in its actuated cage when close enough.

\textbf{Level 4: Shoot.} If the target is a goal and the robot has a ball, it should shoot the ball when it is close enough. 

%The last actuator~$m_5$ is used to open/close the gate. This event-triggered mechanism occurs during ball capture (close gate) and goal scoring (open gate). 

Let~$\ballincage \in \{0, 1 \}$ indicate whether the agent currently has a ball in its cage or not. 

Let~$\ballinsight \in \{0, 1 \}$ indicate whether there is at least one ball in the agent's Field of View (FOV) or not.

Let~$\chi \in \{0, 1 \}$ indicate whether there is at least one goal in the agent's FOV or not. 

The current mode of operation of the agent can then be determined by Algorithm~\ref{alg:perception}.

\subsection{Reduced Agent Model}\label{se:problem}

Here we present a simple kinematic 2.5D model that ignores pitch and roll of our agent. It should be noted that this model is not a good representation of the true system and is only utilized to formally frame our problems and show exactly how we implement their solutions on our real hardware. 

\begin{figure}[!ht]
\centerline{\includegraphics[width=\linewidth]{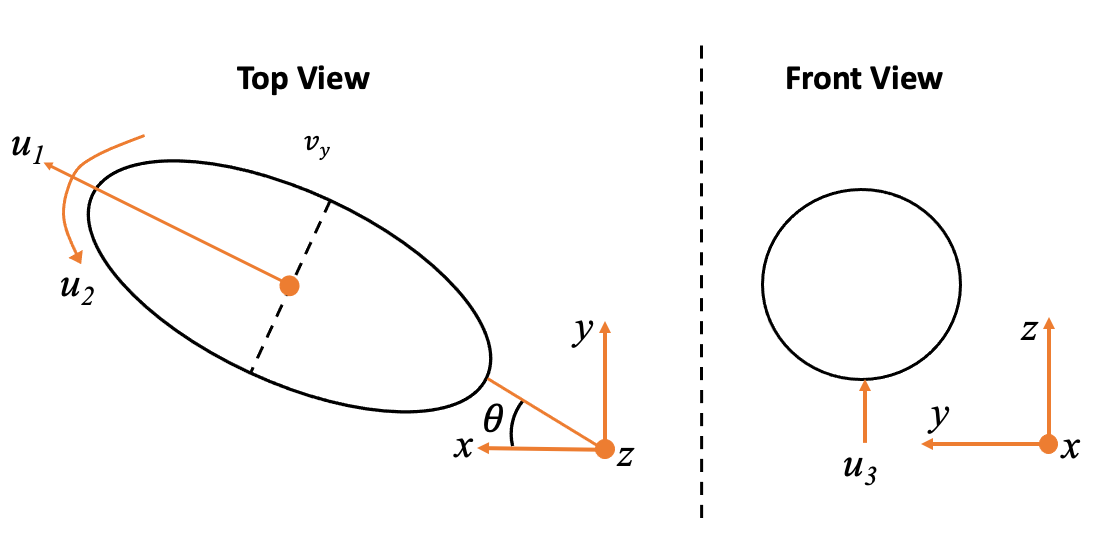}}
\caption{Kinematic model of the robot.}
\label{fig:dynamics}
\end{figure}

The state of the robot is given by its 3D position~$(x,y,z)$ and its 2D orientation~$\theta$. Fig.~\ref{fig:dynamics} shows the simplified model.  Rather than thinking of direct motor inputs, we consider the kinematics
\begin{align}
    \dot{x} &= u_{1} \cos{\theta} \notag \\
    \dot{y} &= u_{1} \sin{\theta}  \label{eq:kinematics} \\
    \dot{\theta} &= u_2 \notag \\
    \dot{z} &= u_3 \notag
\end{align}
where~$u_1,u_2,u_3$ are all bounded by their hardware limits. The immediate coupling of motor~$m_3$ and kinematic input~$u_3$ and $m_4,u_1$ directly controlling the height and forward thrust of the agent are trivial, and the motors~$m_1,m_2$ used as a differential drive command the yaw through~$u_2$. 

In equation~\eqref{eq:kinematics}, $\dot{x},\dot{y},\dot{z}$, and $\dot{\theta}$ is the velocity of the robot in the x-direction, y-direction, z-direction, and angular velocity about the z-axis in a global frame. It should be noted that this global state will \textbf{never} be available to the agents in general and is only used for us to formalize the control problems and present our solutions.

\subsection{Mode Detection and Basic Perception}\label{se:perception}

Here we discuss how the agent estimates what mode it is in and how to process the information needed in each mode. 

Our simple control strategy only requires a minimal level of perception, allowing the agents to know which of the 4 operational modes it should currently be in by estimating the three binary variables~$\gamma,\sigma,\chi$, and where exactly the ball or goal of interest is depending on the mode. 

\begin{figure}
    \centering
    \includegraphics[width=0.8\linewidth]{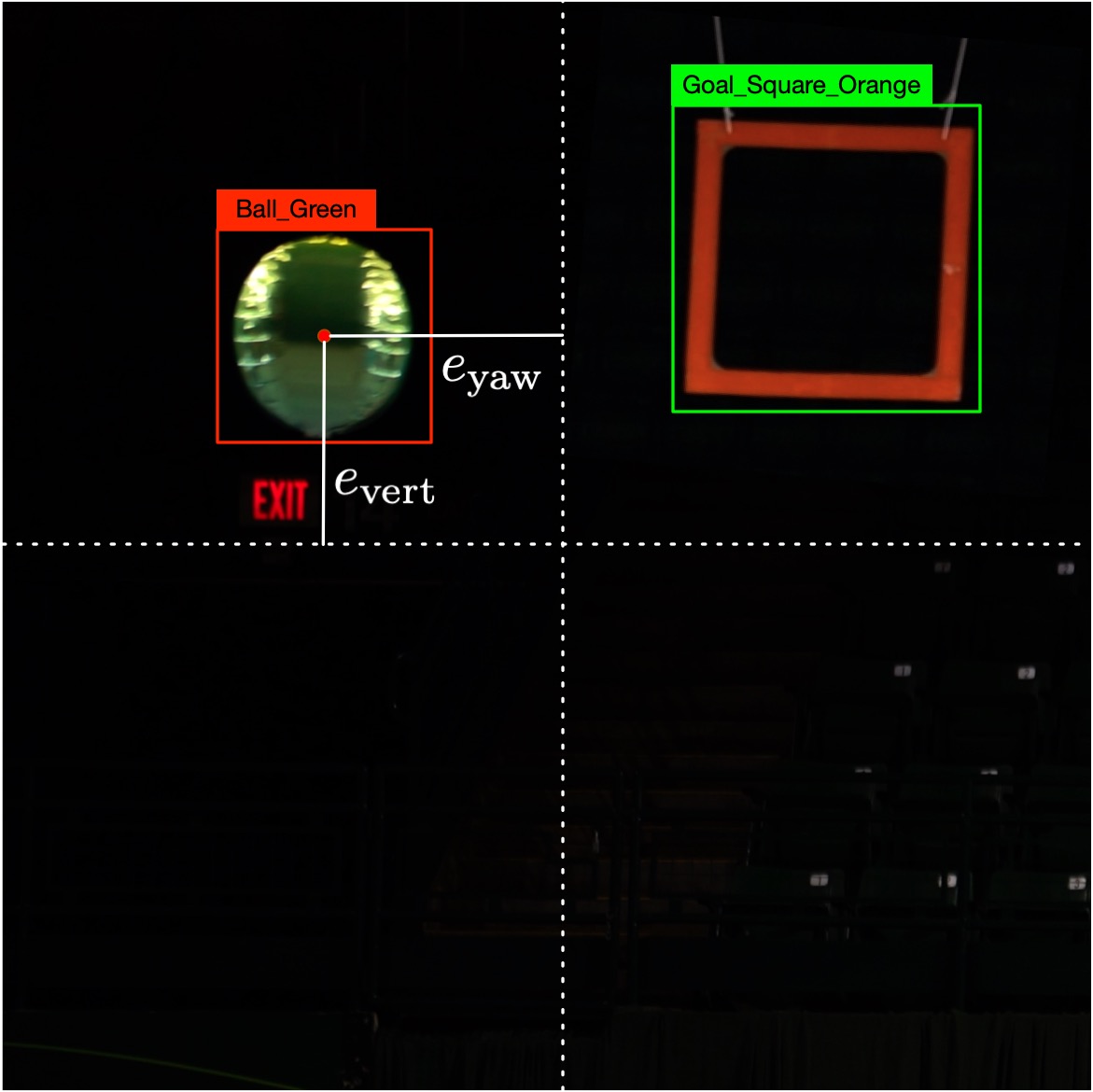}
    \caption{Getting the errors~$e_\text{yaw}$ and~$e_\text{vert}$ that our PD controllers drive to 0 (to effectively center the object of interest in the center of the camera frame). Here we show 2 objects of interest: Green ball and Orange Square Goal.}
    \label{fig:Yolo2PIDErrors}
\end{figure}

\begin{enumerate}
\item \textbf{Have Ball? $\gamma \in \{0,1\}$.} The first thing the agent needs to keep track of is whether it currently is holding a ball in its cage or not. One way to do this using the camera, depending on the position of the camera relative to the cage, is to determine whether the pixels in the camera frame that capture the cage are green or purple (the color of game balls). Simple color detection algorithms can be used to determine the RGB values of specified pixels and comparing them against a tuned threshold~\cite{Fleyeh2004}. However, to have more robust detection in presense of background noise, we deployed a trained yolov5 model~\cite{Redmon2016} to detect balls and its relative size and position in the frame. \begin{archiveText}The details of the detection system is shown in appendix~\ref{se:yolov5}.\end{archiveText}

\item \textbf{See Ball? $\sigma \in \{0,1\}.$} If the robot is not holding a ball~$\gamma = 0$, then it needs to determine if it sees a ball or not. This is as simple as checking whether it sees any green or purple in its video stream. Again here we rely on the yolov5 object detection model to detect the green/purple balls. The training set is tailored to the specific environment so as to enable reliable detection. If a ball is seen~$\sigma = 1$, we need a method of determining both the lateral/yaw offset~$e_\text{yaw}$ and vertical offset~$e_\text{vert}$ from the center of the ball to the center of the camera frame. Note for simplicity we are assuming the center of the camera frame aligns with the center of the cage to properly capture the ball, but depending on the relative position of camera and cage, the offset may be measured from a different point in the camera frame. Fig.~\ref{fig:Yolo2PIDErrors} shows a screenshot of a real camera frame taken from a blimp during a live game and how~$e_\text{yaw},e_\text{vert}$ are measured. \begin{archiveText}Exactly how we do this using the camera is detailed in Appendix~\ref{se:yolov5}.\end{archiveText}

\item \textbf{See Goal? $\chi \in \{0,1\}.$} If the robot is holding a ball~$\gamma = $ then it needs to determine if it sees a goal or not. This is as simple as checking whether it sees any yellow or orange in its video stream. Similarly to when a ball is found, we want to estimate the offsets~$e_\text{raw},e_\text{vert}$. However, estimating these quantities for the goals are more challenging than the balls because they are hollow objects. Fig.~\ref{fig:Yolo2PIDErrors} shows a screenshot of a real frame where~$e_\text{yaw},e_\text{vert}$ must be estimated with respect to the center of the hollow orange objects. In addition to the offsets the size of the goal must also be estimated from the video stream which will be useful when scoring the goal. \begin{archiveText}The details of how we do this are in Appendix~\ref{se:yolov5}.\end{archiveText}
\end{enumerate}

%
%%One possible candidate is to capitalize on the game rules of specific color of the ball and goals; The balls are always a shade of green and the goals are either orange or yellow. Using a camera system to take an image of the game field a color detection algorithm~ can be used as a simple detector. This works very well for a big blob of color in the camera frame. Since a ball is a very regular object, this can be easily tuned for the ball. However, the goal is not a continuous blob; there is a huge hole in the middle and this presents a big challenge for traditional color detection methods. This means simple color-blob detection does indeed meet the minimum viable requirements of our system, but false and missed detection can be quite problematic. 
%
%We discuss in Appendix~\ref{se:yolov5} how to use more advanced software/hardware combination to improve detection performance of the game balls and goals. 

  \begin{algorithm}[t]
                \caption{Ball/Goal Search}
                \label{alg:Search}
                \hspace*{\algorithmicindent}  \textbf{While} $\famode=1$ OR $\famode=3$
                \begin{algorithmic}[1]
                	\If {$\famode=1$ (Ball search) and $~\sigma = 1$ (See ball)}
                	\State gate = Open
                	\Else
                	\State gate = Closed
                	\EndIf
                    \State set~$u_2 = \turnvelsel$ (spin around slowly)
                    \State set~$u_1 = \fwdvelsel$ (move forward while spinning)
                    \State alternate~$u_3 \in \{-\updwnvelsel,\updwnvelsel\}$ to randomly move up and down
                \end{algorithmic}
            \end{algorithm}

  \begin{algorithm}[t]
                \caption{Ball Capture}
                \label{alg:BallCapture}
                \hspace*{\algorithmicindent}  \textbf{While} $\famode=2$
                \begin{algorithmic}[1]
                	\State gate = Open
                    \State set~$u_1 = \fwdvelsel$ (move forward)
                    \State set~$u_2$ PD controller on $e_\text{yaw}$
                    \State set~$u_3$ PD controller on $e_\text{vert}$
                \end{algorithmic}
            \end{algorithm}

  \begin{algorithm}[t]
                \caption{Goal Score}
                \label{alg:GoalScore}
                \hspace*{\algorithmicindent}  \textbf{While} $\famode=3$
                \begin{algorithmic}[1]
                	\State gate = Closed
                    \State set~$u_1 = \fwdvelsel$ (move forward)
                    \State set~$u_2$ PD controller on $e_\text{yaw}$
                    \State set~$u_3$ PD controller on $e_\text{vert}$
                    \If {Goal size greater than threshold size}
                        \State gate = Open
                        \State set $u_1 = -\bar{u}_1$ (blow balls out)
                    \EndIf 
                \end{algorithmic}
            \end{algorithm}

\subsection{Control}\label{se:control}

Thanks to the design of our agent, we have a very simple controller for all 4 modes of operation. 

Intuitively, the input~$u_1$ is generally always set to $\fwdvelsel$, a desired forward velocity to allow the agent to move forward like a Dubin's vehicle. The only time it changes~$u_1 = -\fwdvelsel$ to blow the balls out when scoring a goal.

The input~$u_2$ is used to steer or control the yaw of the robot. When searching for a ball or goal, we simply full throttle~$u_2 = \turnvelsel$ to spin around until it sees something of interest in modes~$\famode = 1,3$. When a target (either ball or goal) is available,~$u_2$ uses a simple PD controller to drive~$e_\text{yaw} \rightarrow 0$. 

The input~$u_3$ is used to control the altitude or height of the robot. When searching for a ball or goal, we simply toggle~$u_3 \in \{-\updwnvelsel,\updwnvelsel\}$ randomly to move up and down using a selected velocity in the 3D environment until it sees something of interest in modes~$\famode=1,3$. When a target is available,~$u_3$ uses a simple PD controller to drive~$e_\text{vert} \rightarrow 0$. 

Finally, the cage gate control is actuated to be open only in modes $\famode = 2, 4$. In $\famode = 2$ the gate remains open all the time facilitating ball capture. In $\famode = 4$, we have the gate open only when the agent is close to a goal. This can be detected when the size of goal being tracked is above a preset threshold that determined through experimentation. 

\begin{figure}[!ht]
\centerline{\includegraphics[width=0.8\linewidth]{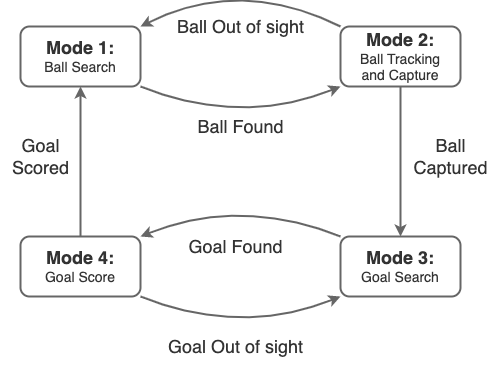}}
\caption{State Machine for Autonomous Behaviors}
\label{State_Machine}
\end{figure}

Coupled with the basic perception system determining the mode of operation~$\famode \in \{1,2,3,4\}$, this autonomous behavior of the robot is described using a simple finite automata with the 4 modes shown in Fig.~\ref{State_Machine}. This is a continuous process until the robot powers off, or a human interrupt is invoked.

\begin{figure}[!ht]
    \centering
    \includegraphics[width=0.8\linewidth]{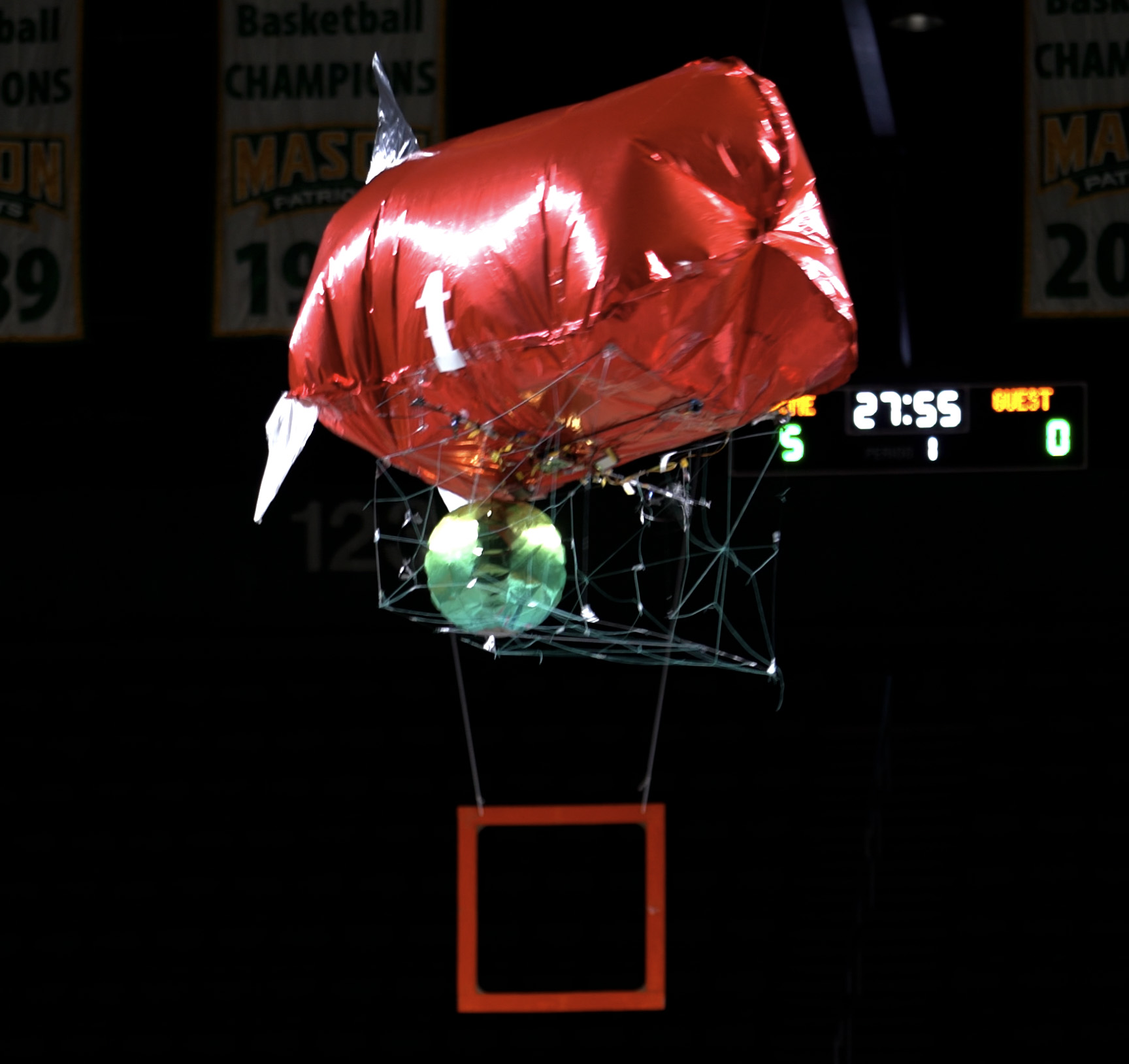}
    \caption{Images of our deployed agent with one captured ball.}
    \label{cage}
\end{figure}

\begin{figure}[!ht]
\centerline{\includegraphics[width=7.5cm,height=7cm]{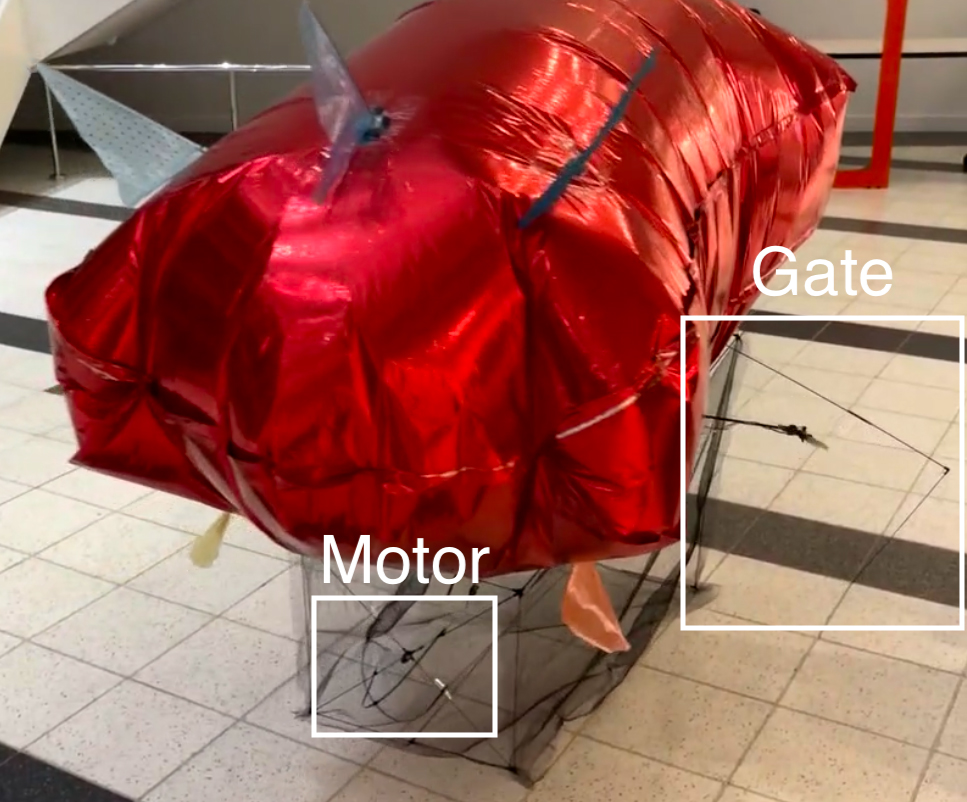}}
\caption{Image showing the motor on the back of the cage and the gate in the front of the cage.}
\label{gate}
\end{figure}

The fully built agent in action is shown in Figs.~\ref{cage} and~\ref{gate}.

\section{Lessons Learned and Future Recommendations}\label{se:upgrades}
    We detail some of the observations we made in the DTR competition conducted in April 2023 at Eagle Bank Arena in George Mason University. 

    \subsection{Gaining an Edge}
    The use of learning based object detection system in our robotic system made detection of objects of interest in the game field very good. This was made possible by the monolithic controller board design that allowed all the complex processing systems to be miniaturized to as small as $72\unit{g}$ of weight. It should be acknowledged that our strong performance was strongly correlated to a change of venue in which there was a lot of background green and yellow colors (as George Mason University hosted this competition for the first time). More specifically, we relied on in-situ labeling and training of objects of interest in the middle of the competition week to greatly improve our perception capabilities.

    %In the past, we struggled detection of goals due to the "hole in the middle" and traditional color based detection (which the game was naturally inclined to use due to the specific setup) was unreliable. With the other teams still using this color based detection, out team had a huge advantage. Coupled with the very simple control algorithm, our system was very effective in the game. 

    \subsection{Limitations and Room for Improvement}

    \subsubsection{System Setup Time}
        The setup time for our entire fleet amounted to more than 2 days. This includes inflating the envelopes, assembling the agents, balancing the agents, testing basic motion and finally performing autonomous capture and score tests. The small idiosyncrasies between each agent mean that we spend a lot of time making minor tweaks so that all agents perform the ball capture and score to a minimal degree. One of the major factors here is the experimental nature of the project where the design of the system is constantly evolving and so there are very few standardized parts. This leads to changing assembly procedures that lead to these differences observed in agents. 
        
Needing to manually tune each agent separately due to the small idiosyncrasies is not scaleable. New approaches for both (i) consistency and speed of designing and deploying agents and (ii) self-tuning methods are desired.

%These sub assemblies can be attached to a skeletal structure using friction mounts thereby making the whole design modular. Having standardizes subassemblies (for example actuation system, where the motor, mount and ESCs are fashioned into a single part using a 3D printed part) also makes repairs more efficient as a new sub assembly can be slapped on the core structure whenever a part breaks during competition. Additionally, pre-made structures make it easier for transportation to different competition arena.

    \subsubsection{Variance in Lighting Condition vs Detection Performance}
        We have observed that the detection performance was heavily dependent on the lighting condition of the arena. Naturally relying on in-situ labeling and training is also not a scalable approach both in terms of setup time but more importantly non-transferrability of robots meaning slight changes in the environment may render algorithms trained earlier to be useless. 

    Using only a single sensor (RGB camera in this case) is a clear limitation. It is desired to integrate other sensing mechanisms to aid in perception, especially in scenarios where the RGB camera is weak. For example, the goals are covered in retro-reflective tape and using an IR light source shining directly onto the goal and using an IR camera for detection makes it much easier to detect from farther away, which may serve as a more course-grained sensor to help the robot get to positions where the camera can do its job properly. 

    \subsubsection{Aerodynamics of the Envelope Design}
    Admittedly one of our most lacking areas is good mechanical engineering design. Controllability, specifically vertical stability (yaw control), was an issue present in all agents. Without properly mechanical engineering analysis and design methods (for instance CFD analysis), we ended up with a much heavier agent than we would like. Propellers are also notoriously inefficient and the efficiency and controllability of our agents should be greatly improved through the use of different modes of propulsion such as control surfaces. 
    
    It is desired to integrate better mechanical engineering practices and aerodynamic trade-offs into our design choices.

\subsubsection{Heterogeneous Fleets}
While our team has explored a lot of different agent designs and even some specialized roles, our team was basically carried in terms of actual point scoring by the single agent design described in this paper. 

The main strategy of our team was to deploy as many agents as possible capable of executing the intended sequence with a non-zero probability. Unfortunately this introduced plenty of issues such as 3 teammates fighting over the same ball due to their low perception capabilities. 

Besides just better coordination among a fleet of homogeneous agents, it is desired to have a heterogeneous team of agents where different agents have specialized roles and can work together to play DTR. 

While our single agent here focused on reliably moving to precise locations to capture balls and score goals with little room for errors, novel methods that can deal with bad perception in other ways should be explored. While our approach relied heavily on simply improving perception in any way possible to allow our PD controllers to drive agents/balls to within a 5-10cm error tolerance, other methods can rely on much larger capturing mechanisms to tolerate much larger errors in perception (e.g., an aerial pursing net~\cite{Julian2019, TokyoNetDrone}).

%
%    \subsection{Future Designs for Task Specific Blimps}
%        The current design is a minimally viable LTA agent capable of capturing balls and scoring goals that was successfully deployed in the recent game. Our current team consisted only of one type of agent. In addition to the improvements mentioned to the backbone agents we are looking to enable more robotic designs that would enable disrupting and maybe incapacitating enemy agents in the field, allow cooperative ball passing strategies and finally sweep all balls in a giant fishing net. These heterogeneous multi-agent system will bring more exciting capabilities to the game. 

\section*{Acknowledgements} 

This work was supported in part by the Department of the Navy, Office of Naval Research (ONR), under federal grants N00014-20-1-2507 and N00014-23-1-2222. 

\bibliographystyle{resources/IEEEtran}
\bibliography{references}

\begin{archiveText}

\appendix

\subsection{Envelope Design}\label{se:envelope}

%The envelope to hold helium was made out of metallized film. In order to have sufficient lift to support all the necessary payload on the agent, a custom envelope was needed. Using metallized film, envelopes were manufactured to hold large volumes of helium. %(not sure the specific type. Asked Fluffy years ago and did not get a definitive answer. Her 28" wide rolls are manufactured by Glenroy, Inc in Wisconsin and they have the material as Metallized Biax Nylon and linear low density polyethylene LLDPE.) 
Red and light blue rolls of film were used to make two sets of agents. 
%The colored side of the film is metallized biax nylon, while the uncolored gray side is the LLDPE. 
Through its heat sealing properties, when the LLDPE sides of the film are in contact and heated, an airtight seal is created. This can be done with traditional irons or even a heat gun. Our agents were manufactured using constant heat roller sealers. The silver and color side of the film will not seal when heated to each other. A self-sealing valve from Anagram was used to allow inflation of the envelopes.  

In past competitions, our envelopes have traditionally been 2-layered. This means only 2 sheets of film were layered on top of each other and a seal was created. The maximum width of our agents was restricted by the width of the roll of film 111.76 cm (44 in). The width of the agents had to be slightly less due to leaving a margin room for sealing and the width of the seal itself. The only way to increase the volume of the envelopes was to increase length, but this posed manufacturing challenges and it hindered the aerodynamics of the agent for flight control (loss of pitch stability). 
To resolve this, envelopes with an additional folded layer were manufactured to increase volume without sacrificing aerodynamic stability. The backbone agent envelopes were 3-layers. To achieve this, one sheet of the film is placed down, with the silver side up. Ensure the sheets are as smooth and flat as possible when taping it to a table. Creases and wrinkles raise the risk of holes and tears when heat sealing over. Placing another sheet with the silver side facing down, we then folded this sheet over itself in half. The fold should be exactly over the center-line of the first sheet and the silver side facing up. Do not tape the folded side down, only the edge. Lastly, place another sheet over the other two sheets, with the silver side facing down. To prevent melting and tearing of the film, a PTFE coated fiberglass fabric sheet is placed over the sheets when sealing. When sketching the outline, ensure the fold is the center-line of the agent. If not, the envelopes faces will not be equally sized when inflated. Seal along the outline, but makes sure to leave a gap for the valve. The folding method takes advantage of both sides of the film. Heating the folded side creates multiple seals, while the colored surface allows for the sides to be separated even when heated. Once the envelope is cut from the excess material, take a valve and insert a thin strip of teflon PTFE. This prevents the valve from sealing shut when sealing it to the envelope. Below are the dimensions of our envelope in cm. (Approximately 630 g of lift was provided). Picture below as well

\begin{figure}
    \centering
    \includegraphics[width=\linewidth]{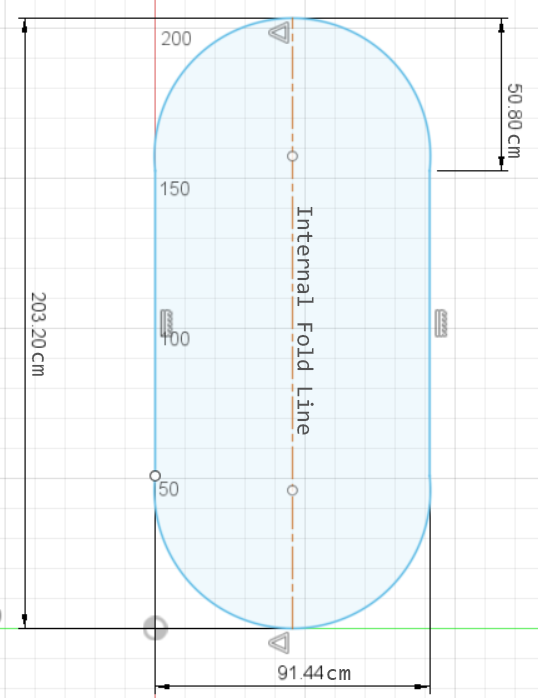}
    \caption{Dimensions used to manufacture the envelope. The internal fold line here shows the inner layer folded along this line to create the 3 layered envelope}
    \label{fig:Envelope_CAD_Dims}
\end{figure}

\subsection{Cage Design}\label{se:cage}

The sides of the cage are made of $3\times 3\times1000$mm carbon fiber rods. The front face of the cage where the target is captured is square in shape and has a side length of 600mm. This gives us an area of $3.6\times 10^5$mm$^2$. The targets are spherical and have a diameter of approximately $500mm$ which gives a cross-section area of $1.96×10^5$mm$^2$. The reason for making the cage's frontal area bigger than the cross-section area of the target is to tolerate impreciseness in the controller to an extent. In Fig.~\ref{cage}, we show the cage frontal area versus the target cross-section area during the capture (left) and that five targets can be captured simultaneously (right). In figure \ref{gate}, we show the motor mounted on the back of the cage and the gate which is used to capture the target.

\subsection{Improving Detection using YOLOv5}
    \label{se:yolov5}
    What we need for goal detection is the ability to incorporate the shape of goals on top of existing color detection methods. More recently we have seen use of learning based object detection frameworks used for detection of complex objects in the visual data. Some of the object detection systems gaining prominence in the last decade are~\cite{Zou2019,Zhao2019,Redmon2016,Girshick2015} with mode advanced image segmentation techniques~\cite{Minaee2022} that allow even 3D pose estimation in some scenarios~\cite{Hu2019}. Our main focus in this paper will be on the popular "You Only Look Once" \cite{Redmon2016} single shot object detection framework and its application as an target detection framework for the LTA agents that will be deployed for DTR.

    To detect objects (balls or goals) we deploy a pre-trained YOLOv5~\cite{Jocher2022} object detection model. A YOLOv5 system has the ability to detect and localize objects of a specific class in an image. This is done by employing a two-step process: first a preliminary bounding box regression is performed followed by predicting the class of object. 
    
    %The high level components of a YOLOv5 network are explained below:
    %\begin{itemize}
    %    \item \textbf{Backbone Network}: YOLOv5 starts with a backbone network, such as Darknet53 or EfficientNet, which extracts features from the input image. This backbone network consists of several layers of convolutional neural networks (CNNs) that progressively learn abstract representations of the image they call an activation map.
    %    \item \textbf{Neck}: After the backbone network, YOLOv5 incorporates a "neck" module, which further refines the extracted features. The neck module typically includes additional convolutional layers, often with a combination of upsampling and downsampling operations, to enhance the spatial and semantic information.
    %    \item \textbf{Detection Head:} The refined features from the neck are then passed through the detection head, which is responsible for predicting bounding boxes and class probabilities. The detection head consists of a set of convolutional layers followed by a final set of fully connected layers.
    %\end{itemize}
    
    YOLOv5 employs anchor boxes (pre-defined boxes of different sizes and aspect ratios) which are placed at various positions in the image and serve as reference points for the network to predict bounding box coordinates, thereby assisting in object localization. In the specific implementation of, YOLOv5, the output layer generates a large number of bounding box proposals by predicting offsets from the anchor box positions. These offsets represent the coordinates for the top-left and bottom-right corners of the predicted bounding boxes. Outside the neural network framework, we use non-maximum suppression (NMS) to eliminate redundant bounding box proposals from the network output and improve detection accuracy. NMS filters out overlapping bounding boxes based on their confidence scores and overlap of bounding boxes and selects the ones with the highest probabilities among clusters of bounding boxes with a certain threshold of overlap. For each remaining bounding box, YOLOv5 predicts the probability distribution across different object classes. This is achieved using softmax activation on the class scores, indicating the likelihood of each class being present within the bounding box.

    \begin{figure}[h]
        \centering
        \includegraphics[width=\linewidth]{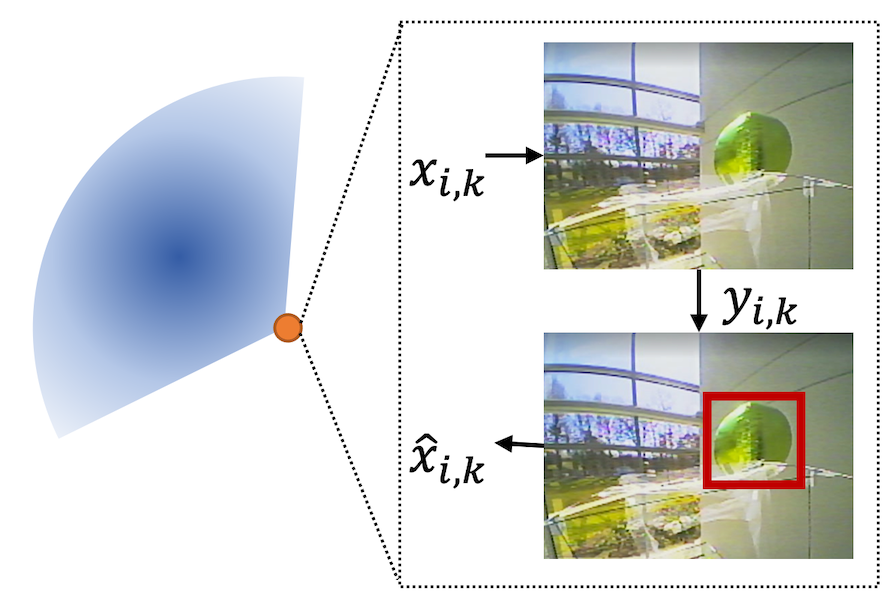}
        \caption{Illustration of the bounding box around the detected target in the FOV of the agent. The bounding box is used to extract the relative position of the target from the agent.}
        \label{fig:ObjectDetectionCamera}
    \end{figure}
    
    In short, for any input image, YOLOv5 outputs three pieces of information for each target $k$ detected: a bounding box around the detected target, the class of the target and the confidence of detection as shown in figure~\ref{fig:ObjectDetectionCamera}. Let the bounding box for the target~$k$ (either a ball or a goal) be~$\hat{y}^{bb}_{k} \in \real^4 := [\hat{y}^{bbx^-}_{k}, \hat{y}^{bbx^+}_{k}, \hat{y}^{bby^-}_{k}, \hat{y}^{bby^+}_{k}]^T$ consisting of the bounds of the box in each orthogonal axes of the frame. We can extract the relative position measurement~$\hat{x}_{k} \in \real^3 := [\hat{x}^r_{k}, \hat{x}^\theta_{k}, \hat{x}^\psi_{k}]^T$ of the object using the bounding box using:
                
    \begin{align}
        \label{eq:targetdetection}
        \hat{x}_k = \begin{bmatrix}
            \sqrt{k^r / \left(({y}^{bbx^+}_{k} - \hat{y}^{bbx^-}_{k}) (\hat{y}^{bby^+}_{k} - \hat{y}^{bby^-}_{k})\right)}\\
            k^\theta ({y}^{bbx^+}_{k} + \hat{y}^{bbx^-}_{k}) / 2 \\
            k^\psi ({y}^{bby^+}_{k} + \hat{y}^{bby^-}_{k}) / 2
        \end{bmatrix}
    \end{align}

    where~$k^\theta$ and~$k^\psi$ is the transformation factor required to change the pixel value of the detected object to bearing angles in the horizontal and vertical axes respectively. $k^r$ is the transformation factor that converts the pixel area to the distance to the target from the camera frame. We obtain this from the calibration experiment for the camera.

    \subsubsection{Training Dataset}
    \label{se:dataset}
    For training the YOLOv5 detection model, we need to create a dataset consisting of all the objects we are interested in identifying. This process of dataset creation uses labelled images. We collect a large number of images containing the target objects and mark out exact bounding boxes of them. This process uses a tool called LabelImg \cite{heartexlab} which is a popular graphical image annotation tool used for labeling and annotating objects in images. We show a sample data labelling session in figure~\ref{fig:LabelImgLabel}
    
    In-situ training seems to have given us the edge over the competition this time.

    \begin{figure}[h]
        \centering
        \includegraphics[width=0.8\linewidth]{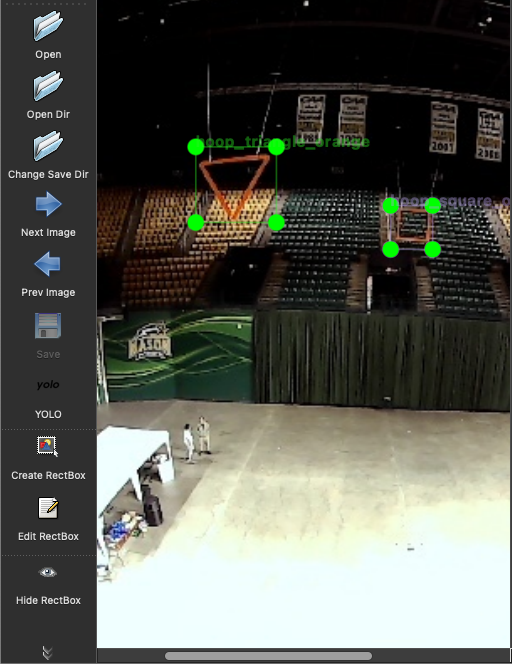}
        \caption{Using labelimg to create labelled datasets for images taken from the Eagle Bank Arena. Here we show an orange square and orange triangle hoops labelled}
        \label{fig:LabelImgLabel}
    \end{figure}

    Our dataset has 8 classes in total consisting of balls (green and purple) and goals (orange and yellow color; square, circle and triangle shape). While creating these datasets we have inferred the following:
    \begin{itemize}
        \item For each class of objects we need at least 200 images and about less than 400 images will do. All images of each class must be unique with very few repetitions of background. It helps a lot to separate out images of each class to different folders for easier labelling.
        \item Images of the objects must include them at various distances from the camera. Some images must be close, even clipped in the FOV, while the others must be far away. This helps with improving detection performance at different distances.
        \item Some objects in the frame might falsely trigger the detection for a specific class. One example of this is the fact that blooming from the bright lights from eagle bank arena were detected as balls as the pattern created in the bloom resembled a stretched out balloon. The way to mitigate this is to create a dataset with these artifacts that trigger false positives without any labelling on these (include at least one object that belongs to the actual class in these images too). During training the regions without a label are considered as background and hence penalize these erroneous detections. 
        \item While labelling make sure all the objects that are visible are labelled with tight tolerance. Adding more background to the object bounding box increases the chance for false detections.
    \end{itemize}

    For training we need to split the training images to test and train datasets. To get the dataset to the final form required by the training, there is a python script that compiles all the images, does a random split based on test train ratio and finally compiles them to the structure required by the YOLOv5 training script.

    \subsubsection{FlySensei: Onboard vision processing system}
    \label{se:FlySensei_implementation}

    \begin{figure}[t]
        \centering
        \includegraphics[width=0.9\linewidth]{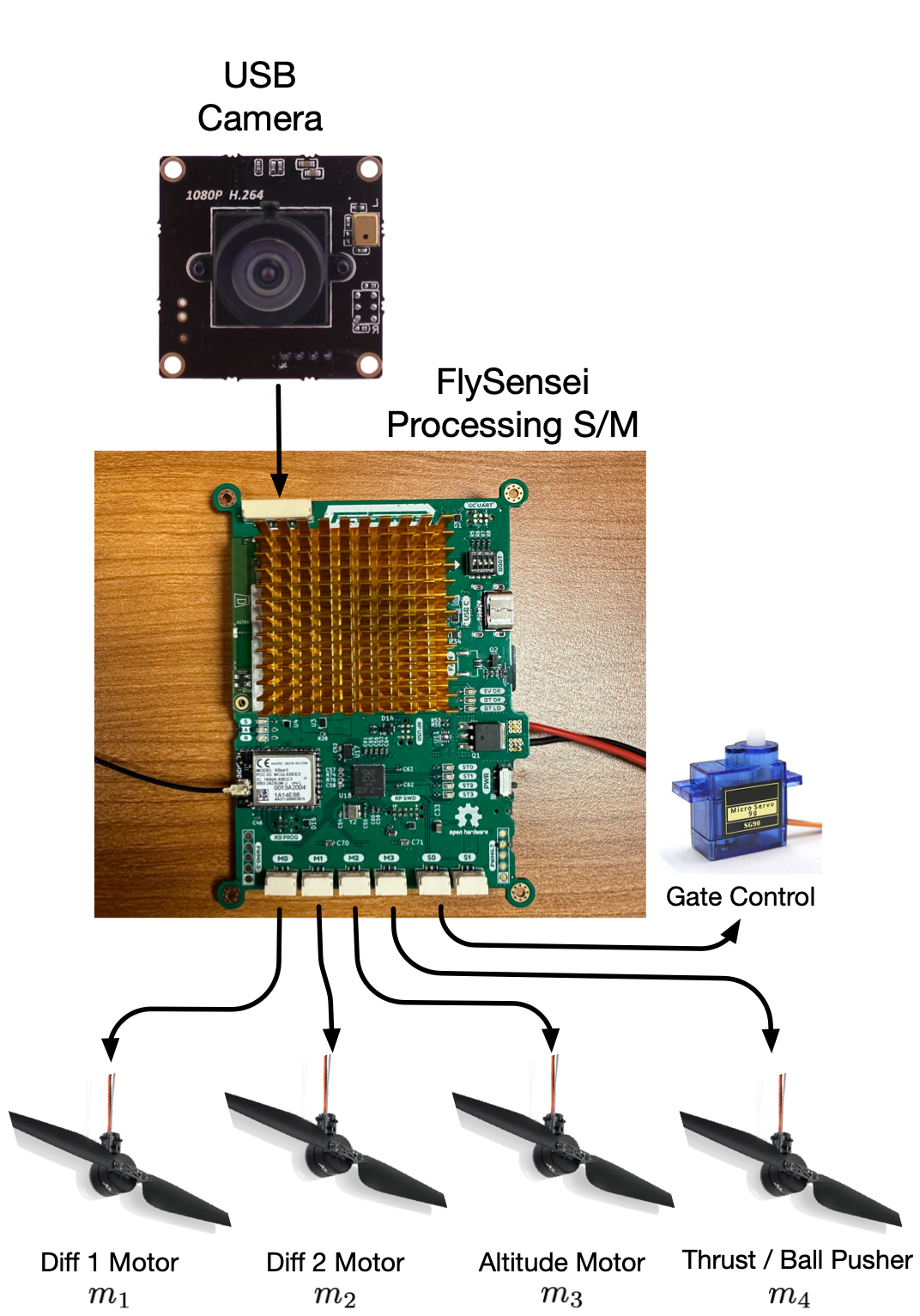}
        \caption{Onboard Processing pipeline for LTA agents}
        \label{fig:SystemImplementation}
    \end{figure}

    We need to implement the detection algorithm on an LTA Agent that can capture images and process them. To that end, we created a fully integrated solution called FlySensei. We show the system in figure~\ref{fig:SystemImplementation}. This system uses a Google Coral SOM, which runs the YOLOv5 model. The vision pipeline starts with a USB camera that is placed in front of the agent. This captured image data is piped to the Google coral. Here we run the YOLOv5 model that outputs the state estimates of the target as mentioned on section~\ref{se:yolov5}. This state information is used for the PID based motor command generation also mentioned in section~\ref{se:yolov5}.

    To get a usable model for YOLOv5 that can be run on the Google Coral SOM, we need to first train the model using the dataset created specifically for the blimps. We found that the use of the following parameters yielded a model that could be used: The input layer size for the network is $320 \times 320 \times 3$. This is an image of $320 \times 320$ pixels. YOLOv5 has multiple model complexities defined and we could go as high as the small model before the model became too big to implement in google coral. The training was conducted to 1000 epochs with a batch size of 32. 

    \begin{figure}[h]
        \centering
        \includegraphics[width=\linewidth]{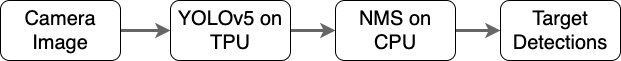}
        \caption{Vision pipeline on Google Coral SOM}
        \label{fig:visionPipeline}
    \end{figure}

    The output model from training cannot be directly run on the Google Coral. The system expects an int8 quantized TF Lite model that can run the network on the TPU onboard. To achieve this we use an existing converter with the best checkpoint of the trained network. To note here is that the neural network does not include the NMS filter that is needed and so we have to implement a version of that to run on the CPU side of the coral SOM. The full image detection pipeline is shown in figure~\ref{fig:visionPipeline}.

\end{archiveText}

\end{document}